\documentclass{article}

\usepackage{graphicx}
\usepackage{psfrag}
\usepackage{subfigure}
\usepackage{algorithm}
\usepackage{algorithmic}
\usepackage{amsmath}
\usepackage{amssymb}
\usepackage{amsthm}
\usepackage{bbm}
\usepackage[normalem]{ulem}
\usepackage{hyperref}
\usepackage{breakurl}

\newcommand{\R}{\mathbb{R}}
\newcommand{\Order}{\mathcal{O}}
\newcommand{\sign}{\operatorname{sign}}
\newcommand{\fp}[2]{\frac{\partial #1}{\partial #2}}
\newcommand{\TT}[1]{\mbox{\tiny $#1$}}

\newcommand{\bmat}{\begin{pmatrix}}
\newcommand{\emat}{\end{pmatrix}}

\begin{document}

\title{Accelerated Linear SVM Training with\\Adaptive Variable Selection Frequencies\\[2em]}

\author{Tobias Glasmachers\\
		\texttt{tobias.glasmachers@ini.rub.de}\\
		Institut f\"ur Neuroinformatik,\\
        Ruhr-Universit\"at Bochum, 44780 Bochum, Germany\\[1ex]
		\"Ur\"un Dogan\\
		\texttt{doganudb@math.uni-potsdam.de}\\
		Institut f\"ur Mathematik,\\
		Universit\"at Potsdam, Germany\\[2em]}

\date{}

\maketitle

\begin{abstract}
Support vector machine (SVM) training is an active research area since
the dawn of the method. In recent years there has been increasing
interest in specialized solvers for the important case of linear models.
The algorithm presented by Hsieh et al., probably best known under the
name of the ``liblinear'' implementation, marks a major breakthrough.
The method is analog to established dual decomposition algorithms for
training of non-linear SVMs, but with greatly reduced computational
complexity per update step. This comes at the cost of not keeping track
of the gradient of the objective any more, which excludes the
application of highly developed working set selection algorithms. We
present an algorithmic improvement to this method. We replace uniform
working set selection with an online adaptation of selection
frequencies. The adaptation criterion is inspired by modern second order
working set selection methods. The same mechanism replaces the shrinking
heuristic. This novel technique speeds up training in some cases by more
than an order of magnitude.
\end{abstract}

\pagebreak

%%%%%%%%%%%%%%%%%%%%%%%%%%%%%%%%%%%%%%%%%%%%%%%%%%%%%%%%%%%%%%%%%%%%%%%%
\section{Introduction}
\label{sec:introduction}

Since the pioneering work of Joachims \cite{joachims:1998} and
Platt~\cite{platt:1998}, support vector machine (SVM) training is
dominated by decomposition algorithms solving the dual problem. This
approach is at the core of the extremely popular software libsvm,
although a number of important algorithmic improvements have been added
over the years~\cite{fan:2005}.

For linear SVM training, the situation is different. It has been
observed that a direct representation of the primal weight vector is
computationally advantageous over optimization of dual variables, since
the dimensionality of the optimization problem is independent of the
number of training patterns. At first glance this hints at solving the
primal problem directly, despite the non-differential nature of the
hinge loss. Influential examples of this research direction are the
cutting plane approach \cite{joachims:2006} and the stochastic
gradient descent ``Pegasos'' algorithm \cite{shalev:2007}.

Hsieh et al.~\cite{hsieh:2008} were the first to notice that it is
possible to solve the dual problem with a decomposition algorithm
similar to those used for non-linear SVM training while profiting from
the fixed dimensionality of the weight vector. This method combines fast
(linear) convergence of its kernelized counterpart and a direct
representation of the weight vector, which allows to perform update
steps in time independent of the size of the training set. The method
has been demonstrated to outperform several algorithms for direct
optimization of the primal problem \cite{hsieh:2008}. We refer the
reader to the excellent review \cite{yuan:2012} and references therein
for a detailed discussion of the differences, as well as for the
relation of the method to non-linear (kernelized) SVM training.

Our study builds upon this work. We present an algorithmic improvement
of the dual method \cite{hsieh:2008}. This new method differs from
the existing algorithm in two aspects, namely the selection of the
currently active sub-problem and the shrinking heuristic. The selection
of the working set, defining the sub-problem in the decomposition
algorithm, has been subject to extensive research
\cite{keerthi:2002,fan:2005,glasmachers:2006}. However, these elaborate
methods are not affordable in the algorithm \cite{hsieh:2008}, and they
are replaced with systematic sweeps over all variables. We propose a
more elaborate method that takes recent experience into account and
adapts selection frequencies of individual variables accordingly. As a
side effect, this algorithm replaces the existing shrinking heuristic
\cite{joachims:1998,hsieh:2008}. Our experimental evaluation shows that
this new method can achieve considerable speed-ups of more than an order
of magnitude.

The remainder of this paper is organized as follows. In the next section
we review the dual training algorithm by \cite{hsieh:2008} and introduce
our notation. Then we present our modifications in
section~\ref{sec:working-set-selection} and an extensive experimental
evaluation thereof in section~\ref{sec:experiments}. We discuss the
results (section~\ref{sec:discussion}) and close with our conclusions
(section~\ref{sec:conclusion}).

%%%%%%%%%%%%%%%%%%%%%%%%%%%%%%%%%%%%%%%%%%%%%%%%%%%%%%%%%%%%%%%%%%%%%%%%
\section{Linear SVM Training in the Dual}
\label{sec:liblinear}

In this section we describe the dual algorithm for linear SVM training
\cite{hsieh:2008}, as implemented in the software
liblinear~\cite{liblinear}. It should be noted that the liblinear
software supports a large number of training methods for linear models,
such as binary and multi-category classification and regression, as well
as different regularizers and loss functions. In this study we restrict
ourselves to the most basic case, which is the ``standard'' SVM with
hinge loss and two-norm regularizer. However, our proceeding is general
in nature and therefore applicable to many of the above cases.

Given a binary classification problem described by training data
\begin{align*}
	\big( (x_1, y_1), \dots, (x_\ell, y_\ell) \big) \in (\R^d \times \{-1, +1\})^\ell
	\enspace,
\end{align*}
SVM training corresponds to finding the solution of the (primal)
optimization problem
\begin{align*}
	\min_{w \in \R^d} \quad \frac{1}{2} \|w\|^2 + C \cdot \sum_{i=1}^\ell L(y_i, \langle w, x_i \rangle)
	\enspace,
\end{align*}
where $L(y, f(x)) = \max\{0, 1 - y \cdot f(x)\}$ is the hinge loss and
$C > 0$ controls the solution complexity
\cite{cortes:1995,schoelkopf:2002}. The prediction of the machine is of
the form $h : \R^d \to \{-1, +1\}$, $h(x) = \sign(f(x))$, based on the
linear decision function $f(x) = \langle w, x \rangle$.
In this formulation we have dropped the constant offset $b$ that is
often added to the decision function, since it turns out to be of minor
importance in high dimensional feature spaces, and dropping the term
results in attractive algorithmic simplifications
(see, e.g,\cite{steinwart:2011}).

The corresponding dual optimization problem is the box constrained
quadratic program
\begin{align*}
	\max_{\alpha \in \R^\ell} \quad & W(\alpha) = \sum_{i=1}^\ell \alpha_i - \frac{1}{2} \sum_{i,j=1}^\ell \alpha_i \alpha_j y_i y_j \langle x_i, x_j \rangle \\
	\text{s.t.} \quad & 0 \leq \alpha_i \leq C \quad \forall \, i \in \{1, \dots, \ell\}
	\enspace.
\end{align*}
It holds $w = \sum_{i=1}^\ell y_i \alpha_i x_i$. Since the dual training
method is rooted in non-linear SVM training, we want to mention that in
general all inner products $\langle x_i, x_j \rangle$ between training
examples are replaced with a usually non-linear Mercer kernel
function~$k(x_i, x_j)$.

A standard method for support vector machine training is to solve the
dual problem with a decomposition algorithm \cite{osuna:1997,joachims:1998,bottou:2007}.
The algorithm decomposes the full quadratic program into a sequence of
sub-problems restricted to few variables. The sub-problems are solved
iteratively until an overall solution of sufficient accuracy is found.
Sequential minimal optimization (SMO,~\cite{platt:1998}) refers to
the important special case of choosing the number of variables in each
sub-problem minimal. For the above dual this minimal number is one, so
that the algorithm essentially performs coordinate ascent. The skeleton
of this method is shown in algorithm~\ref{algo:SMO}.

\begin{algorithm}
\begin{algorithmic}
	\REPEAT 
		\STATE select active variable $i$
		\STATE solve sub-problem restricted to variable $i$
		\STATE update $\alpha_i$ and further state variables
	\UNTIL{(all KKT violation $< \varepsilon$)}
	\caption{SMO without equality constraint}
	\label{algo:SMO}
\end{algorithmic}
\end{algorithm}

The training algorithm for linear SVMs by \cite{hsieh:2008}
is an adaptation of this technique. Its crucial algorithmic improvement
over standard SMO is to reduce the complexity of each iteration from
$\Order(\ell \cdot d)$ to only $\Order(d)$.%
\footnote{The algorithm is particularly efficient for sparse inputs.
  Then the complexity of $\Order(d)$ is further reduced to $\Order(nnz)$,
  where $nnz$ is the number of non-zero components of the currently
  selected training example~$x_i$.}
The key trick is to keep track of the primal vector $w$ during the
optimization. This allows to rewrite the derivative of the dual
objective function as
\begin{align*}
	\fp{W}{\alpha_i} = 1 - \sum_{j=1}^\ell \alpha_j y_i y_j \langle x_i, x_j \rangle
			= 1 - \alpha_i y_i \langle x_i, w \rangle
	\enspace,
\end{align*}
which can be computed in $\Order(d)$ operations. The requirement to
perform all steps inside the SMO loop in $\Order(d)$ operations makes
some changes necessary, as compared to standard SMO. For instance, the
flat SMO loop is split into an outer and an inner loop. The full
algorithm is provided in detail as algorithm~\ref{algo:liblinear}.

\begin{algorithm}
\begin{algorithmic}
	\STATE $A \leftarrow \{1, \dots, \ell\}$; $v_{\min}^\text{old} \leftarrow -\infty$; $v_{\max}^\text{old} \leftarrow \infty$
	\LOOP
		\STATE $v_{\min} \leftarrow \infty$; $v_{\max} \leftarrow -\infty$
		\FORALL{$i \in A$ in random order}
			\STATE $g_i \leftarrow 1 - y_i \cdot \langle x_i, w \rangle$
			\IF{$\alpha_i = 0$ and $g_i < v_{\min}^\text{old}$}
				\STATE $A \leftarrow A \setminus \{i\}$
			\ELSIF{$\alpha_i = C$ and $g_i > v_{\max}^\text{old}$}
				\STATE $A \leftarrow A \setminus \{i\}$
			\ELSE
				\STATE \textbf{if} ($\alpha_i > 0$ and $g_i < v_{\min}$) \textbf{then} $v_{\min} \leftarrow g_i$
				\STATE \textbf{if} ($\alpha_i < C$ and $g_i > v_{\max}$) \textbf{then} $v_{\max} \leftarrow g_i$
				\STATE $\mu \leftarrow \left[ g_i / \|x_i\|^2 \right]_{-\alpha_i}^{C-\alpha_i}$
				\STATE $\alpha_i \leftarrow \alpha_i + \mu$
				\STATE $w \leftarrow w + \mu \cdot y_i \cdot x_i$
			\ENDIF
		\ENDFOR
		\IF{($v_{\max} - v_{\min} < \varepsilon$)}
			\STATE \textbf{if} ($A = \{1, \dots, \ell\}$) \textbf{then} \textbf{break}
			\STATE $A \leftarrow \{1, \dots, \ell\}$; $v_{\min}^\text{old} \leftarrow -\infty$; $v_{\max}^\text{old} \leftarrow \infty$
		\ELSE
			\STATE \textbf{if} $v_{\min} < 0$ \textbf{then} $v_{\min}^\text{old} \leftarrow v_{\min}$ \textbf{else} $v_{\min}^\text{old} \leftarrow -\infty$
			\STATE \textbf{if} $v_{\max} > 0$ \textbf{then} $v_{\max}^\text{old} \leftarrow v_{\max}$ \textbf{else} $v_{\max}^\text{old} \leftarrow \infty$
		\ENDIF
	\ENDLOOP
	\caption{``liblinear'' algorithm}
	\label{algo:liblinear}
\end{algorithmic}
\end{algorithm}

Most prominently, the selection of the active variable
$i \in \{1, \dots, \ell\}$ (defining the sub-problem to be solved in the
current iteration) cannot be done with elaborate heuristics that are key
to fast training of non-linear SVMs
\cite{keerthi:2002,fan:2005,glasmachers:2006}. Instead, the algorithm
performs systematic sweeps over all variables. The order of variables is
randomized. The (amortized) complexity of selecting the active variable
is $\Order(1)$.

The solution of the sub-problem amounts to
\begin{align*}
	\alpha_i \leftarrow \left[ \alpha_i + \frac{1}{\|x_i\|^2} \frac{\partial W}{\partial \alpha_i} \right]_0^C
		= \left[ \alpha_i + \frac{1 - y_i \langle x_i, w \rangle}{\|x_i\|^2} \right]_0^C
	\enspace,
\end{align*}
where $[x]_a^b = \max\{a, \min\{b, x\}\}$ denotes clipping to the
interval $[a, b]$. This operation, requiring two inner products, is done
in $\Order(d)$ operations (in the implementation the squared norm is
precomputed).

The usual SMO proceeding to keep track of the dual gradient
$\nabla_\alpha W(\alpha)$ is not possible within the tight budget of
$\Order(d)$ operations. Instead the weight vector is updated. Let
$\mu = \alpha_i^\text{new} - \alpha_i^\text{old}$ denote the step
performed for the solution of the sub-problem, then the weight update
reads $w \leftarrow w + \mu \cdot y_i \cdot x_i$, which is an
$\Order(d)$ operation.

The usual stopping criterion is to check the maximal violation of the
Karush-Kuhn-Tucker (KKT) optimality conditions. For the dual problem
the KKT violation can be expressed independently for each variable. Let
$g_i = \fp{W}{\alpha_i}$ denote the derivative of the dual objective.
Then the violation is $|g_i|$ if $0 < \alpha_i < C$, $\max\{0, g_i\}$
for $\alpha_i = 0$, and $\max\{0, -g_i\}$ for $\alpha_i = C$. The
algorithm is stopped as soon as this violations drops below some
threshold~$\varepsilon$ (e.g., $\varepsilon = 0.001$).

In the original SMO algorithm this check is a cheap by-product of the
selection of the index~$i$. Since keeping track of the dual gradient is
impossible, the exact check becomes an $\Order(\ell \cdot d)$ operation.
This is the complexity of a whole sweep over the data. In
algorithm~\ref{algo:liblinear} the exact check is therefore replaced
with an approximate check, where each variable is checked at the time it
is active during the sweep. Thus, all variables are checked, but not
exactly at the time of stopping. The algorithm keeps track of
$v_{\min} = \min\{ g_i \,|\, \alpha_i > 0 \}$ and
$v_{\max} = \max\{ g_i \,|\, \alpha_i < C \}$, and check for
$v_{\max} - v_{\min} < \varepsilon$ at the end of the sweep.

To exploit the sparsity of the SVM solution, the algorithm is equipped
with a shrinking heuristic. This heuristic removes a variable from the
set $A$ of active variables if it is at the bounds and the gradient of
the dual objective function indicates that it will stay there. After a
while, this heuristic can remove most variables from the problem, making
sweeps over the variables much faster. The drawback of this heuristic is
that it can fail. Therefore, at the end of the optimization run, the
algorithm needs to check optimality of the deactivated variables. The
detection of a mistake results in continuation of the loop, which can be
costly.
Therefore, the decision to remove a variable needs to be conservative.
The algorithm removes a variable only if it is at a bound and its
gradient $g_i$ pushes against the bound with a strength that exceeds the
maximal KKT violation of active variables during the previous sweep.
This amounts to the condition $g_i < v_{\min}^\text{old}$ for
$\alpha_i = 0$ and to $g_i > v_{\max}^\text{old}$ for $\alpha_i = C$.

%%%%%%%%%%%%%%%%%%%%%%%%%%%%%%%%%%%%%%%%%%%%%%%%%%%%%%%%%%%%%%%%%%%%%%%%
\section{Online Adaptation of Variable Selection Frequencies}
\label{sec:working-set-selection}

In this section we introduce our algorithmic improvement to the above
described linear SVM training algorithm. Our modification targets two
weaknesses of the algorithm at once.
\begin{itemize}
\item
	Algorithm~\ref{algo:liblinear} executes uniform sweeps over all
	active variables. In contrast to the SMO algorithm for non-linear
	SVM training, the selection is not based on a promise of the
	progress due to this choice. Although the computational restriction
	of $\Order(d)$ operations does not allow for a search for the best
	a-priori guarantee of some sort (such as the largest KKT violation),
	we can still learn from the observed progress after a step has been
	executed.
\item
	Shrinking of variables is inevitably a heuristic.
	Algorithm~\ref{algo:liblinear} makes ``hard'' shrinking decisions by
	removing variables based on adaptive thresholds on the strength with
	which they press against their active constraints. It is problematic
	that even a single wrong decision to remove a variable early on can
	invalidate a large share of the algorithm's (fine tuning) efforts
	later on. Therefore we replace this mechanism with what we think of
	as ``soft'' shrinking, which amounts to the reduction of the
	selection frequency of a variable, down to a predefined minimum.
\end{itemize}

In algorithm~\ref{algo:liblinear} there are only two possible
frequencies with which variables are selected. All inactive variables
are selected with frequency zero, and all active variables are selected
with the same frequency $1/|A|$. This scheme is most probably not
optimal; it is instead expected that some variables should be selected
much more frequently than others.

Established working set selection heuristics aim to pick the best (in
some sense) variable for the very next step, and therefore automatically
adapt relative frequencies of variable selection over time. This is not
possible within the given framework. However, we can still use the
information of whether a step has made good progress or not to adapt
selection frequencies for the future. This adaptation process is similar
to so-called self-adaptation heuristics found in modern direct search
methods, see e.g.~\cite{hansen:2001}. To summarize,
although we are unable to determine the best variable for the present
step, we can still use current progress as an indicator for future
utility.

For turning this insight into an algorithm we introduce adaptive
variable selection frequencies based on preference values $p_i > 0$.
The relative frequency of variable $\alpha_i$ is defined as
\begin{align*}
	\frac{p_i}{\sum\limits_{j=1}^\ell p_j}
	\enspace.
\end{align*}
In each iteration of the outer loop the algorithm composes a schedule
(a list of $\ell$ variables indices) that reflects these relative
frequencies. This task is performed by algorithm~\ref{algo:schedule}.
With $\Order(\ell)$ operations it is about as cheap as the randomization
of the order of variables.
We call this novel variable selection scheme \emph{adaptive variable
selection frequencies} (AVSF).

The crucial question is: how to update the preferences $p_i$ over the
course of the optimization run? For this purpose the gain
$\Delta = W(\alpha^\text{new}) - W(\alpha^\text{old})$ of an iteration
with active variable $\alpha_i$ is compared to the average (reference)
gain $\Delta_\text{ref}$. Since the average gain decreases over time,
this value is estimated as a fading average. The preference is changed
by the rule
\begin{align*}
	p_i \leftarrow \Big[ p_i \cdot \exp \big( c \cdot (\Delta - \Delta_\text{ref}) \big) \Big]_{p_{\min}}^{p_{\max}}
	\enspace.
\end{align*}
In our experiments we set the constants to $c = 1/5$,
$p_{\min} = 1/20$, and $p_{\max} = 20$. The bounds
$0 < p_{\min} \leq p_{\max} < \infty$ ensure that the linear convergence
guarantee established by Theorem~1 in \cite{hsieh:2008} directly
carries over to our modified version. The adaptation of preference
values is taken care of by algorithm~\ref{algo:preferences}. The added
complexity per iteration is only $\Order(1)$.

The dual objective gain $\Delta$ is used in modern second order working
set selection algorithms for non-linear SVM training
\cite{fan:2005,glasmachers:2006}. Our method resembles this highly
efficient approach; it can be understood as a time averaged variant.

It is important to note that the above scheme does not only increase the
preferences and therefore the relative frequencies of variables that
have performed above average in the past. It also penalizes variables
that do not move at all, typically because they are at the bounds and
should be removed from the active set: such steps give the worst
possible gain of zero. Thus, the algorithm quickly drives their
preferences to the minimum. However, they are not removed completely
from the active set. Checking these variables from time to time is a
good thing, because it is cheap compared to uniform sweeps on the one
hand and at the same time avoids that early mistakes are discovered only
very much later.

Another difference to the original algorithm is that shrinking decisions
are not based on KKT violations, but instead on relative progress in
terms of the dual objective function. We are not aware of an existing
approach of this type.

Algorithm~\ref{algo:AVSF} incorporates our modifications into the
liblinear algorithm. The new algorithm is no more complex than
algorithm~\ref{algo:liblinear}, and it requires only a hand full of
changes to the existing liblinear implementation.

\begin{algorithm}
\begin{algorithmic}
	\STATE $p \leftarrow (1, \dots, 1) \in \R^\ell$; $p_\text{sum} \leftarrow \ell$
	\STATE $\Delta_\text{ref} \leftarrow 0$
	\STATE canstop $\leftarrow$ true
	\LOOP
		\STATE $v \leftarrow 0$
		\STATE define schedule $I \in \{1, \dots, \ell\}^\ell$ (algorithm~\ref{algo:schedule})
		\FORALL{$i \in I$ in random order}
			\STATE $g_i \leftarrow 1 - y_i \cdot \langle x_i, w \rangle$
			\STATE \textbf{if} ($\alpha_i > 0$ and $-g_i > v$) \textbf{then} $v \leftarrow -g_i$
			\STATE \textbf{if} ($\alpha_i < C$ and $g_i > v$) \textbf{then} $v \leftarrow g_i$
			\STATE $\mu \leftarrow \left[ g_i / \|x_i\|^2 \right]_{-\alpha_i}^{C-\alpha_i}$
			\STATE $\alpha_i \leftarrow \alpha_i + \mu$
			\STATE $w \leftarrow w + \mu \cdot y_i \cdot x_i$
			\STATE update preferences (algorithm~\ref{algo:preferences})
		\ENDFOR
		\IF{$v < \varepsilon$}
			\STATE \textbf{if} canstop \textbf{then} \textbf{break}
			\STATE $p \leftarrow (1, \dots, 1) \in \R^\ell$; $p_\text{sum} \leftarrow \ell$
			\STATE canstop $\leftarrow$ true
		\ELSE
			\STATE canstop $\leftarrow$ false
		\ENDIF
	\ENDLOOP
	\caption{Linear SVM algorithm with adaptive variable selection frequencies (AVSF)}
	\label{algo:AVSF}
\end{algorithmic}
\end{algorithm}

\begin{algorithm}
\begin{algorithmic}
	\STATE $N \leftarrow p_\text{sum}$
	\STATE $j \leftarrow 0$
	\FOR{$i \in \{1, \dots, \ell\}$}
		\STATE $m \leftarrow p_i \cdot (\ell - j) / N$
		\STATE $n \leftarrow \lfloor m \rfloor$
		\STATE with probability $m - n$: $n \leftarrow n + 1$
		\FOR {$k \in \{1, \dots, n\}$}
			\STATE $I_j \leftarrow i$
			\STATE $j \leftarrow j + 1$
		\ENDFOR
		\STATE $N \leftarrow N - p_i$
	\ENDFOR
	\caption{Definition of the schedule $I$}
	\label{algo:schedule}
\end{algorithmic}
\end{algorithm}

\begin{algorithm}
\begin{algorithmic}
	\STATE $\Delta \leftarrow \mu \cdot (g_i - \mu/2 \cdot \|x_i\|^2)$
	\IF {first sweep}
		\STATE $\Delta_\text{ref} \leftarrow \Delta_\text{ref} + \Delta / \ell$
	\ELSE
		\STATE $h \leftarrow c \cdot (\Delta / \Delta_\text{ref} - 1)$
		\STATE $p_\text{new} \leftarrow \left[ e^h \cdot p_i \right]_{p_{\min}}^{p_{\max}}$
		\STATE $p_\text{sum} \leftarrow p_\text{sum} + p_\text{new} - p_i$
		\STATE $p_i \leftarrow p_\text{new}$
		\STATE $\Delta_\text{ref} \leftarrow (1 - 1 / \ell) \Delta_\text{ref} + \Delta / \ell$
	\ENDIF
	\caption{Update of the preferences $p$}
	\label{algo:preferences}
\end{algorithmic}
\end{algorithm}

%%%%%%%%%%%%%%%%%%%%%%%%%%%%%%%%%%%%%%%%%%%%%%%%%%%%%%%%%%%%%%%%%%%%%%%%
\section{Experimental Evaluation}
\label{sec:experiments}

We compare our adaptive variable frequency selection
algorithm~\ref{algo:AVSF} (AVSF) to the baseline
algorithm~\ref{algo:liblinear} in an empirical study.
For a fair comparison we have implemented our modifications directly
into the latest version of liblinear (version 1.92 at the time of
writing).

The aim of the experiments is to demonstrate the superior training
speed of algorithm~\ref{algo:AVSF} over a wide range of problems and
experimental settings. Therefore we have added time measurement and a
step counter to both algorithms.%
\footnote{The timer measures the runtime of the core optimization loop.
  In particular, data loading is excluded.}

The liblinear software comes with a hard-coded limit of $1000$ outer loop
iterations. We have removed this ``feature'' for the sake of comparison.
Instead we stop only based on the heuristic stopping criterion described
in section~\ref{sec:liblinear}, which is the exact same for both
algorithms. We use the liblinear default of $\varepsilon = 0.01$ as well
as the libsvm default of $\varepsilon = 0.001$ in all experiments.

We ran both algorithms on a number of benchmark problems. In our
comparison we rely on medium to extremely large binary classification
problems, downloadable from the libsvm data website:
\begin{center}
	\url{http://www.csie.ntu.edu.tw/~cjlin/libsvmtools/datasets/}
\end{center}
Table~\ref{tab:datasets} lists descriptive statistics of the data sets.

\begin{table}
\begin{center}
	\begin{tabular}{l|r|r}
		Problem & Instances $(\ell)$ & Features $(d)$ \\
		\hline
		cover~type    & $581,012$ & $54$ \\
		kkd-a         & $8,407,752$ & $20,216,830$ \\
		kkd-b         & $19,264,097$ & $29,890,095$ \\
		news~20       & $19,996$ & $1,355,191$ \\
		rcv1          & $20,242$ & $47,236$ \\
		url           & $2,396,130$ & $3,231,961$ \\
	\end{tabular}
	\caption{ \label{tab:datasets}
		Number of training examples and number of features of the data
		sets used in our comparison.
	}
\end{center}
\end{table}

Test accuracies are of no relevance for our comparison, since both
algorithm deliver the same solution. Only the runtime matters.
Comparing training times in a fair way is non-trivial. This is because
the selection of a good value of the regularization parameter $C$
requires several runs with different settings, often performed in a
cross validation manner. Therefore the computational cost of finding a
good value of $C$ can easily exceed that of training the final model,
and even a good range for $C$ is hard to guess without prior knowledge.
To circumvent this pitfall we have decided to report training times for
a whole range of values, namely
$C \in \{0.01, 0.1, 1, 10, 100, 1000\}$.
We have averaged the timings for the data sets news~20 and rcv1 over
$10$ independent runs in order to obtain stable results.

Our primary performance metric is wall clock time. A related and easier
to measure quantity is the number of update steps. For both algorithms
the complexity of an update steps is $\Order(d)$ computations. Also, the
wall clock time per step is roughly comparable, but we have to note that
a step of the AVSF algorithm is slightly more costly than for the
original algorithm.

\begin{figure*}
	\psfrag{x-3}[c][c]{$10^{-3}$}
	\psfrag{x-2}[c][c]{$10^{-2}$}
	\psfrag{x-1}[c][c]{$10^{-1}$}
	\psfrag{x0}[c][c]{$10^{0}$}
	\psfrag{x1}[c][c]{$10^{+1}$}
	\psfrag{x2}[c][c]{$10^{+2}$}
	\psfrag{x3}[c][c]{$10^{+3}$}
	\psfrag{y-3}[r][l]{$10^{-3}$}
	\psfrag{y-2}[r][l]{$10^{-2}$}
	\psfrag{y-1}[r][l]{$10^{-1}$}
	\psfrag{y0}[r][l]{$10^{+0}$}
	\psfrag{y1}[r][l]{$10^{+1}$}
	\psfrag{y2}[r][l]{$10^{+2}$}
	\psfrag{y3}[r][l]{$10^{+3}$}
	\psfrag{y4}[r][l]{$10^{+4}$}
	\psfrag{y5}[r][l]{$10^{+5}$}
	\psfrag{y6}[r][l]{$10^{+6}$}
	\psfrag{y7}[r][l]{$10^{+7}$}
	\psfrag{C}[l][l]{$C$}
	\psfrag{seconds}[c][c]{seconds}
	\psfrag{cover type}[c][c]{\textbf{\large cover type}}
	\psfrag{kdd-a}[c][c]{\textbf{\large kdd-a}}
	\psfrag{kdd-b}[c][c]{\textbf{\large kdd-b}}
	\psfrag{news-20}[c][c]{\textbf{\large news-20}}
	\psfrag{rcv1}[c][c]{\textbf{\large rcv1}}
	\psfrag{url}[c][c]{\textbf{\large url}}
	\includegraphics[width=0.49\textwidth]{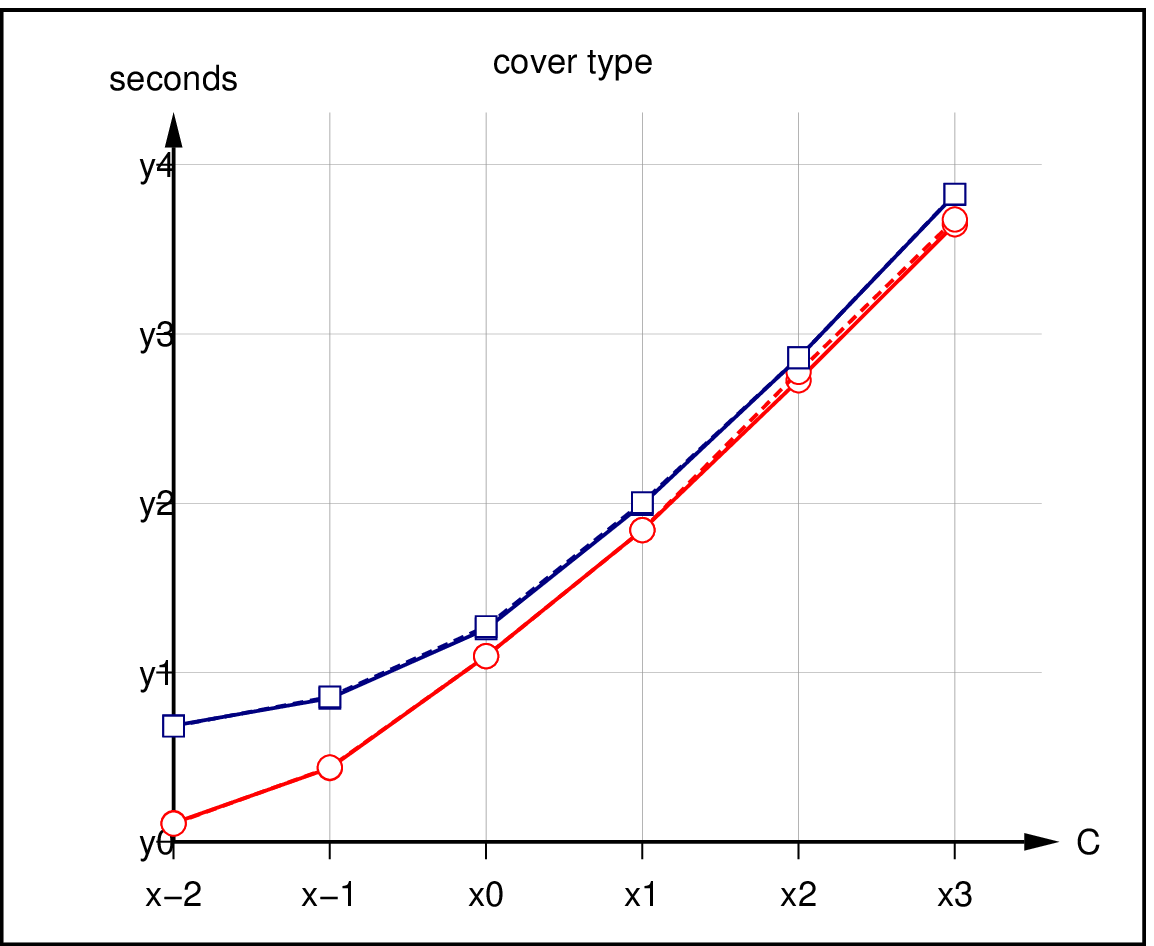}
	~~
	\includegraphics[width=0.49\textwidth]{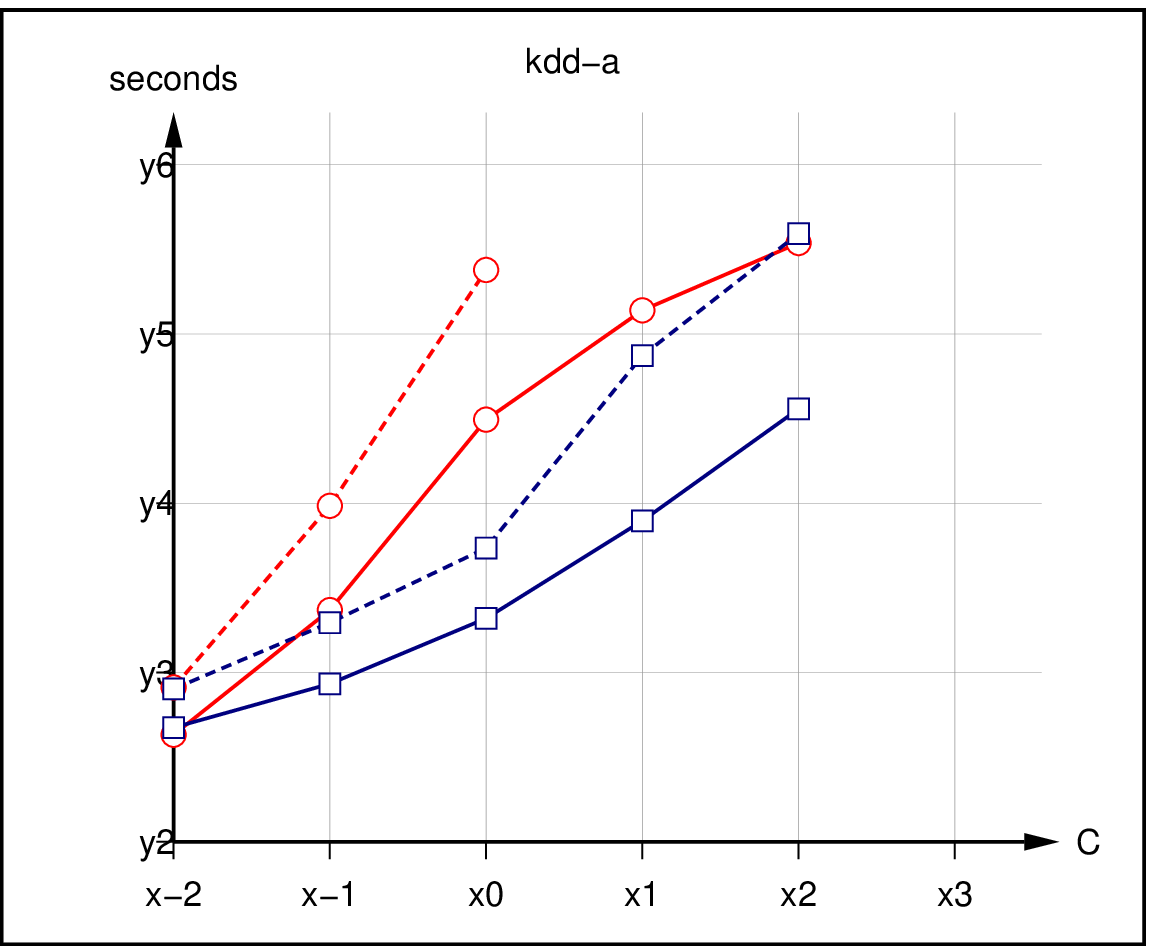}
	\\[1mm]
	\includegraphics[width=0.49\textwidth]{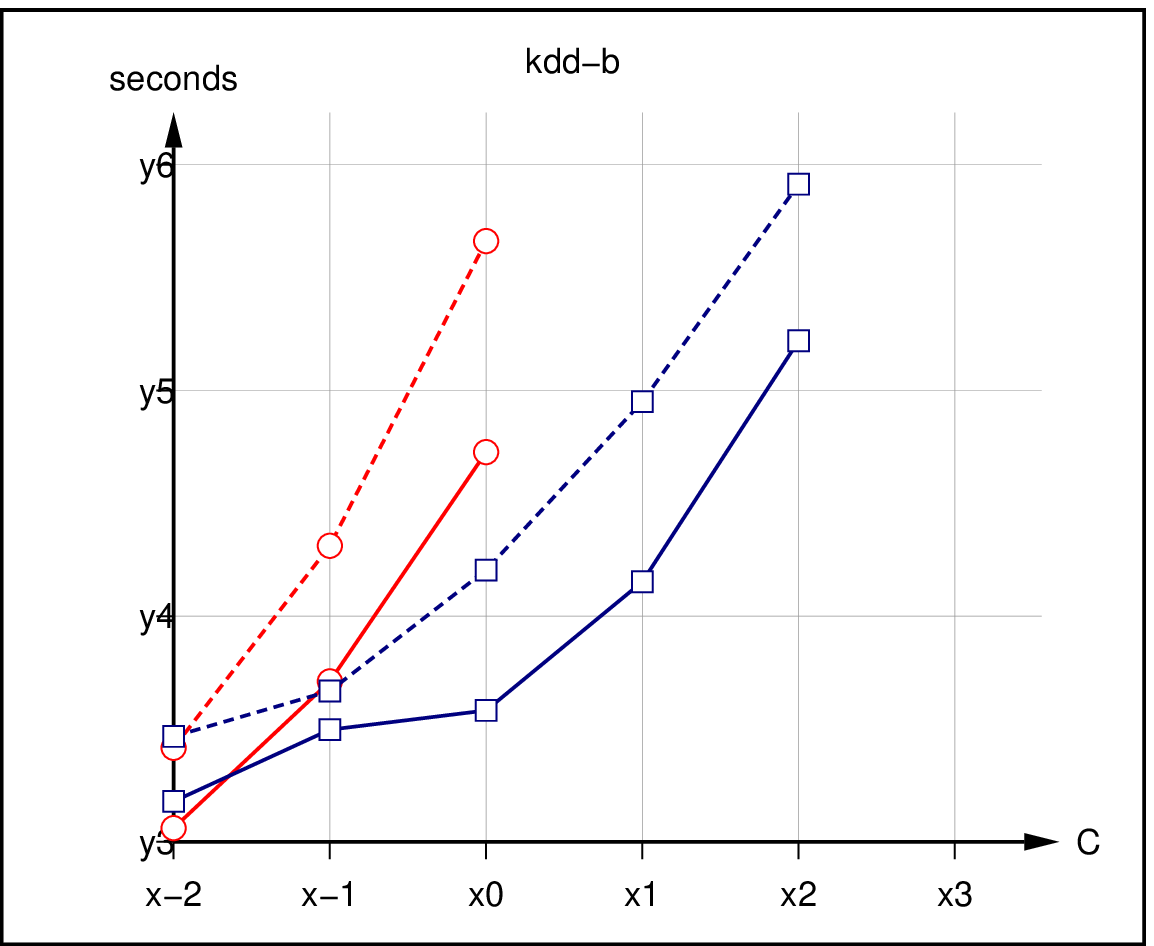}
	~~
	\includegraphics[width=0.49\textwidth]{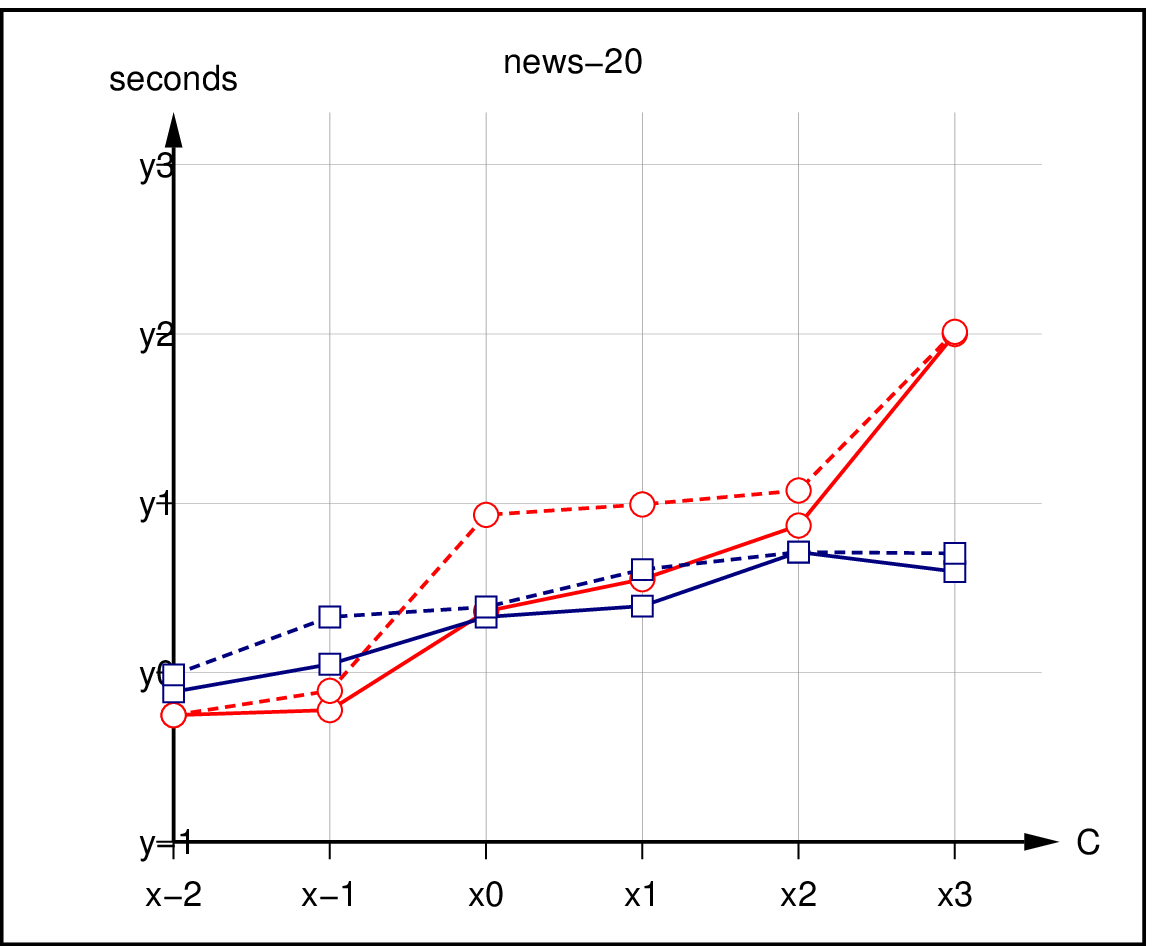}
	\\[1mm]
	\includegraphics[width=0.49\textwidth]{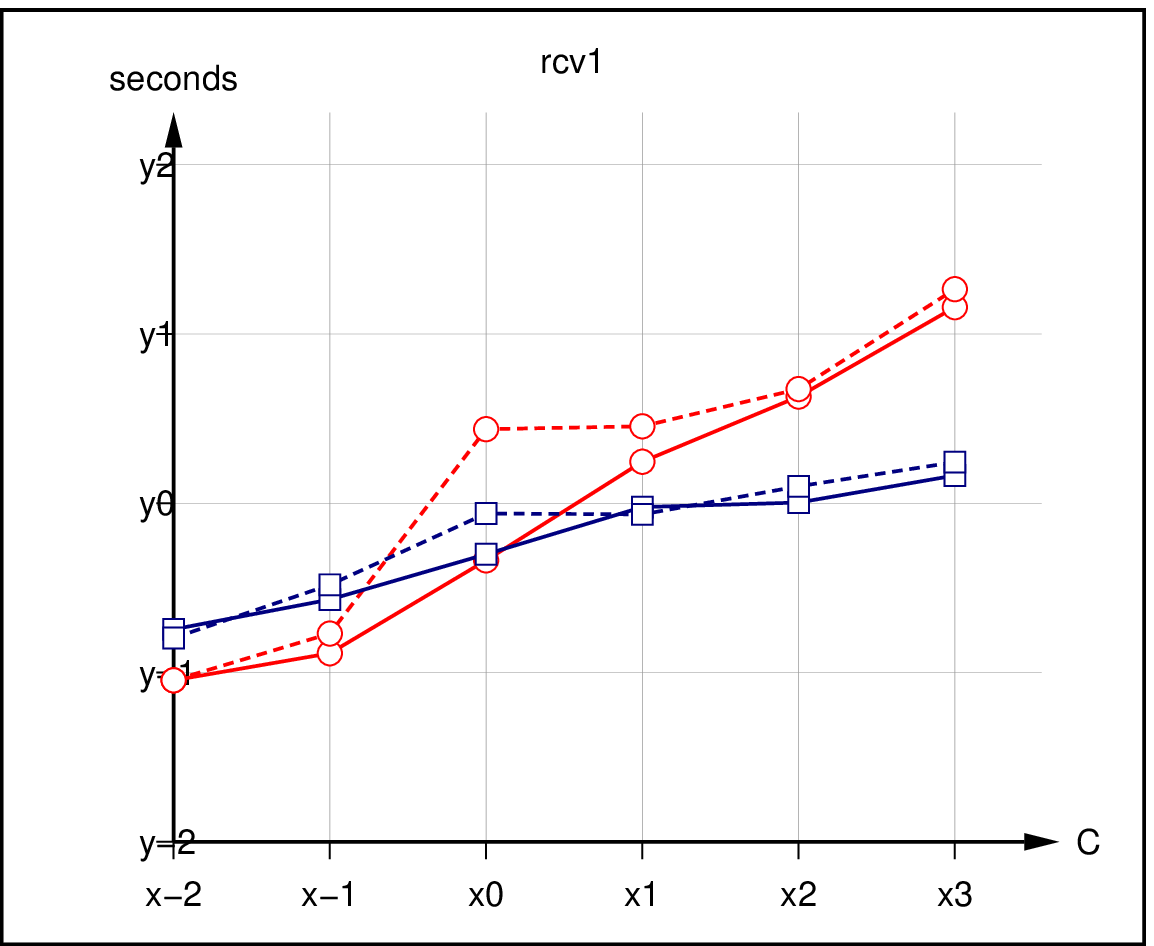}
	~~
	\includegraphics[width=0.49\textwidth]{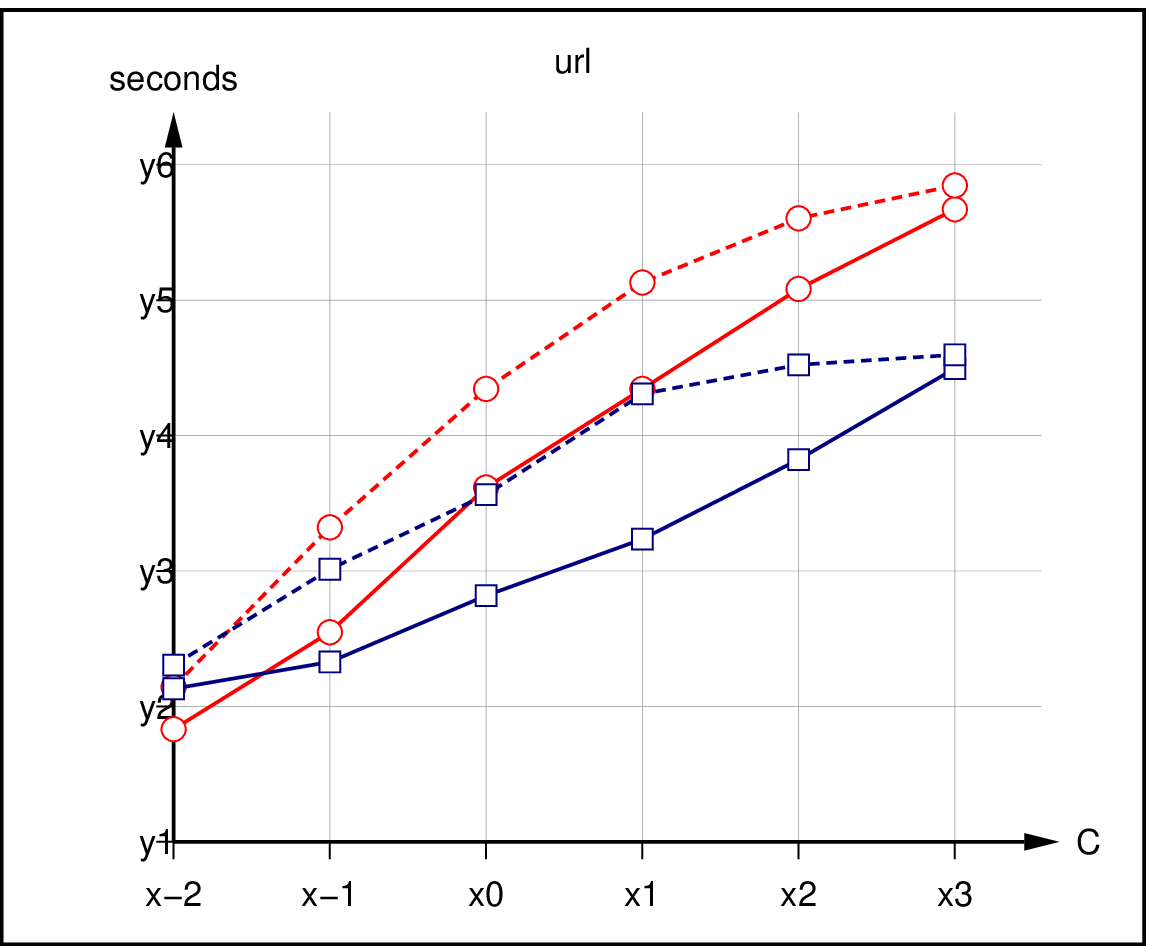}
\begin{center}
	\caption{ \label{fig:results}
		Training times with the original liblinear algorithm
		(red circles) and with our adaptive variable selection algorithm
		(blue squares), as a function of the regularization
		parameter~$C$. The target accuracy is $\varepsilon = 0.01$ for
		the solid curve and $\varepsilon = 0.001$ for the dashed curve.
	}
\end{center}
\end{figure*}

Training times and numbers of iterations are reported in tables
\ref{tab:results1} and \ref{tab:results2}. For an easier comparison,
they are illustrated graphically in figure~\ref{fig:results}.

\begin{table*}
\begin{center}
	\small
	\begin{tabular}{l|c|rrrrrr}
		Problem & Solver & $C=0.01$ & $ C=0.1$ & $C=1$ & $C=10$ & $C=100$ & $C=1000$ \\
		\hline
		\hline
		cover~type    & baseline & $1.29$ & $2.73$ & $12.5$ & $69.5$ & $533$ & $4,450$ \\
		              &          & \TT{3.31 \cdot 10^{6~}} & \TT{7.41 \cdot 10^{6~}} & \TT{3.38 \cdot 10^{7~}} & \TT{1.80 \cdot 10^{8~}} & \TT{1.37 \cdot 10^{9~}} & \TT{1.14 \cdot 10^{10}} \\
		              & AVSF     & $4.86$ & $7.05$ & $18.1$ & $98.4$ & $724$ & $6,670$ \\
		              &          & \TT{8.72 \cdot 10^{6~}} & \TT{1.28 \cdot 10^{7~}} & \TT{3.43 \cdot 10^{7~}} & \TT{1.88 \cdot 10^{8~}} & \TT{1.50 \cdot 10^{9~}} & \TT{1.40 \cdot 10^{10}} \\
		\hline
		kkd-a         & baseline & $429$ & $2,340$ & $31,200$ & $138,000$ & $345,000$ & --- \\
		              &          & \TT{3.07 \cdot 10^{8~}} & \TT{1.57 \cdot 10^{9~}} & \TT{1.88 \cdot 10^{10}} & \TT{8.77 \cdot 10^{10}} & \TT{2.35 \cdot 10^{11}} &        \\
		              & AVSF     & $473$ &   $858$ & $2,090$ & $7,880$ & $36,100$ & --- \\
		              &          & \TT{3.62 \cdot 10^{8~}} & \TT{6.39 \cdot 10^{8~}} & \TT{1.58 \cdot 10^{9~}} & \TT{6.15 \cdot 10^{9~}} & \TT{5.01 \cdot 10^{10}} &        \\
		\hline
		kkd-b         & baseline & $1,150$ & $5,140$ & $53,300$ & $400,000^*$ & $932,000^*$ & --- \\
		              &          & \TT{6.92 \cdot 10^{8~}} & \TT{2.86 \cdot 10^{9~}} & \TT{3.11 \cdot 10^{10}} &        &        &        \\
		              & AVSF     & $1,510$ & $3,140$ & $3,820$ & $14,200$ & $166,000$ & --- \\
		              &          & \TT{7.90 \cdot 10^{8~}} & \TT{1.52 \cdot 10^{9~}} & \TT{2.64 \cdot 10^{9~}} & \TT{7.22 \cdot 10^{9~}} & \TT{8.32 \cdot 10^{10}} &        \\
		\hline
		news~20       & baseline & $0.56$ & $0.60$ & $2.30$ & $3.56$ & $7.39$ & $100$ \\
		              &          & \TT{8.03 \cdot 10^{4~}} & \TT{1.22 \cdot 10^{5~}} & \TT{4.04 \cdot 10^{5~}} & \TT{6.38 \cdot 10^{5~}} & \TT{1.38 \cdot 10^{6~}} & \TT{2.47 \cdot 10^{7~}} \\
		              & AVSF     & $0.77$ & $1.12$ & $2.13$ & $2.47$ & $5.15$ & $3.95$ \\
		              &          & \TT{1.20 \cdot 10^{5~}} & \TT{2.60 \cdot 10^{5~}} & \TT{4.80 \cdot 10^{5~}} & \TT{4.40 \cdot 10^{5~}} & \TT{8.80 \cdot 10^{5~}} & \TT{8.20 \cdot 10^{5~}} \\
		\hline
		rcv1          & baseline & $0.09$ & $0.13$ & $0.46$ & $1.76$ & $4.27$ & $14.1$ \\
		              &          & \TT{9.36 \cdot 10^{4~}} & \TT{1.46 \cdot 10^{5~}} & \TT{4.77 \cdot 10^{5~}} & \TT{1.70 \cdot 10^{6~}} & \TT{4.19 \cdot 10^{6~}} & \TT{1.43 \cdot 10^{7~}} \\
		              & AVSF     & $0.18$ & $0.27$ & $0.50$ & $0.95$ & $1.01$ & $1.46$ \\
		              &          & \TT{1.62 \cdot 10^{5~}} & \TT{2.83 \cdot 10^{5~}} & \TT{4.86 \cdot 10^{5~}} & \TT{9.72 \cdot 10^{5~}} & \TT{1.07 \cdot 10^{6~}} & \TT{1.54 \cdot 10^{6~}} \\
		\hline
		url           & baseline & $67.9$ & $353$ & $4,140$ & $22,100$ & $121,000$ & $469,000$ \\
		              &          & \TT{4.05 \cdot 10^{7~}} & \TT{1.93 \cdot 10^{8~}} & \TT{2.22 \cdot 10^{9~}} & \TT{1.45 \cdot 10^{10}} & \TT{8.04 \cdot 10^{10}} & \TT{2.74 \cdot 10^{11}} \\
		              & AVSF     & $135$ & $213$ & $658$ & $1,720$ & $6,650$ & $31,300$ \\
		              &          & \TT{6.47 \cdot 10^{7~}} & \TT{1.39 \cdot 10^{8~}} & \TT{4.17 \cdot 10^{8~}} & \TT{1.18 \cdot 10^{9~}} & \TT{4.34 \cdot 10^{9~}} & \TT{1.73 \cdot 10^{10}} \\
		\hline
	\end{tabular}
	\caption{ \label{tab:results1}
		Runtime in seconds and number of update steps (inner loop
		iterations, tiny font numbers in scientific notation below) for
		both methods, trained for a range of values of~$C$, with
		target accuracy $\varepsilon = 0.01$.
		Algorithm~\ref{algo:liblinear} is marked with ``baseline'', the
		adaptive variable selection frequencies algorithm~\ref{algo:AVSF}
		with ``AVSF''.
		Runs marked with ``---'' did not finish until the deadline.
		For cases where one algorithm has finished but the competitor
		has not we report the running time until present as a lower
		bound on the true runtime---these entries are marked with a
		star. We want to remark that the actual values may be much
		bigger.
	}
\end{center}
\end{table*}

\begin{table*}
\begin{center}
	\small
	\begin{tabular}{l|c|rrrrrr}
		Problem & Solver & $C=0.01$ & $ C=0.1$ & $C=1$ & $C=10$ & $C=100$ & $C=1000$ \\
		\hline
		\hline
		cover~type    & baseline & $1.28$ & $2.75$ & $12.5$ & $69.5$ & $597$ & $4,750$ \\
		              &          & \TT{3.31 \cot 10^{6~}} & \TT{7.41 \cdot 10^{6~}} & \TT{3.38 \cdot 10^{7~}} & \TT{1.80 \cdot 10^{8~}} & \TT{1.78 \cdot 10^{9~}} & \TT{1.44 \cdot 10^{10}} \\
		              & AVSF     & $4.82$ & $7.18$ & $18.7$ & $101$ & $724$ & $6,710$ \\
		              &          & \TT{8.72 \cdot 10^{6~}} & \TT{1.28 \cdot 10^{7~}} & \TT{3.43 \cdot 10^{7~}} & \TT{1.88 \cdot 10^{8~}} & \TT{1.50 \cdot 10^{9~}} & \TT{1.40 \cdot 10^{10}} \\
		\hline
		kkd-a         & baseline &      $817$ &    $9,660$ &  $239,000$ & $1,800,000^*$ & $1,800,000^*$ & --- \\
		              &          & \TT{1.11 \cdot 10^{9~}} & \TT{9.16 \cdot 10^{9~}} & \TT{1.59 \cdot 10^{11}} &                         &                         &                         \\
		              & AVSF     &      $801$ &    $1,970$ &    $5,440$ &   $74,500$ &  $392,000$ & --- \\
		              &          & \TT{6.22 \cdot 10^{8~}} & \TT{9.84 \cdot 10^{8~}} & \TT{4.23 \cdot 10^{9~}} & \TT{3.80 \cdot 10^{10}} & \TT{2.80 \cdot 10^{11}} &                         \\
		\hline
		kdd-b         & baseline &    $2,610$ &   $20,500$ &  $459,000$ & $1,450,000^*$ & $2,160,000^*$ &        --- \\
		              &          & \TT{1.94 \cdot 10^{9~}} & \TT{1.17 \cdot 10^{10}} & \TT{2.73 \cdot 10^{11}} &                         &                         &                         \\
		              & AVSF     &    $2,930$ &    $4,660$ &   $16,000$ &   $89,200$ &  $820,000$ &        --- \\
		              &          & \TT{1.23 \cdot 10^{9~}} & \TT{2.04 \cdot 10^{9~}} & \TT{7.57 \cdot 10^{9~}} & \TT{4.05 \cdot 10^{10}} & \TT{4.09 \cdot 10^{11}} &                         \\
		\hline
		news~20       & baseline &     $0.56$ &     $0.78$ &     $8.54$ &     $9.84$ &     $11.9$ &      $103$ \\
		              &          & \TT{8.03 \cdot 10^{4~}} & \TT{1.54 \cdot 10^{5~}} & \TT{1.55 \cdot 10^{6~}} & \TT{1.87 \cdot 10^{6~}} & \TT{2.90 \cdot 10^{6~}} & \TT{2.50 \cdot 10^{7~}} \\
		              & AVSF     &     $0.97$ &     $2.13$ &     $2.44$ &     $4.06$ &     $5.15$ &     $6.20$ \\
		              &          & \TT{1.60 \cdot 10^{5~}} & \TT{3.80 \cdot 10^{5~}} & \TT{5.20 \cdot 10^{5~}} & \TT{7.20 \cdot 10^{5~}} & \TT{8.80 \cdot 10^{5~}} & \TT{1.02 \cdot 10^{6~}} \\
		\hline
		rcv1          & baseline &     $0.09$ &     $0.17$ &     $2.74$ &     $2.85$ &     $4.73$ &     $18.4$ \\
		              &          & \TT{9.40 \cdot 10^{4~}} & \TT{1.93 \cdot 10^{5~}} & \TT{3.36 \cdot 10^{6~}} & \TT{3.36 \cdot 10^{6~}} & \TT{5.63 \cdot 10^{6~}} & \TT{2.14 \cdot 10^{7~}} \\
		              & AVSF     &     $0.16$ &     $0.33$ &     $0.87$ &     $0.86$ &     $1.26$ &     $1.75$ \\
		              &          & \TT{1.82 \cdot 10^{5~}} & \TT{3.85 \cdot 10^{5~}} & \TT{1.01 \cdot 10^{6~}} & \TT{9.92 \cdot 10^{5~}} & \TT{1.48 \cdot 10^{6~}} & \TT{2.04 \cdot 10^{6~}} \\
		\hline
		url           & baseline &      $139$ &    $2,100$ &   $22,100$ &  $135,000$ &  $402,000$ &  $703,000$ \\
		              &          & \TT{8.27 \cdot 10^{7~}} & \TT{1.18 \cdot 10^{9~}} & \TT{1.46 \cdot 10^{10}} & \TT{7.61 \cdot 10^{10}} & \TT{2.35 \cdot 10^{11}} & \TT{3.78 \cdot 10^{11}} \\
		              & AVSF     &      $202$ &     $1030$ &    $3,660$ &   $20,300$ &   $33,300$ &   $39,500$ \\
		              &          & \TT{9.82 \cdot 10^{7~}} & \TT{5.65 \cdot 10^{8~}} & \TT{2.48 \cdot 10^{9~}} & \TT{1.00 \cdot 10^{10}} & \TT{2.37 \cdot 10^{10}} & \TT{2.27 \cdot 10^{10}} \\
		\hline
	\end{tabular}
	\caption{ \label{tab:results2}
		Runtime in seconds and number of update steps (inner loop
		iterations, tiny font numbers in scientific notation below) for
		both methods, trained for a range of values of~$C$, with
		target accuracy $\varepsilon = 0.001$.
		Algorithm~\ref{algo:liblinear} is marked with ``baseline'', the
		adaptive variable selection frequencies algorithm~\ref{algo:AVSF}
		with ``AVSF''.
		Runs marked with ``---'' did not finish until the deadline.
		For cases where one algorithm has finished but the competitor
		has not we report the running time until present as a lower
		bound on the true runtime---these entries are marked with a
		star. We want to remark that the actual values may be much
		bigger.
	}
\end{center}
\end{table*}

%%%%%%%%%%%%%%%%%%%%%%%%%%%%%%%%%%%%%%%%%%%%%%%%%%%%%%%%%%%%%%%%%%%%%%%%
\section{Discussion}
\label{sec:discussion}

The typical behavior of the performance timing curves in
figure~\ref{fig:results} is that the original algorithm is superior for
small values of $C$, and that our new algorithm is a lot faster for
large values of $C$: often by an order of magnitude, sometimes even
more. The differences are most pronounced for large data sets.
In the following we will discuss this behavior.

For small values of $C$ many examples tend to become support vectors.
Many dual variables end up at the maximum value. This relatively simple
solution structure can be achieved efficiently with uniform sweeps
through the data. Moreover, the algorithm performs over few outer loop
iterations. In this case our soft shrinking method is too slow to be
effective and the original hard shrinking heuristic has an advantage.
At the same time the problem structure is ``simple enough'', so that
falsely shrinking variables out of the problem is improbable.

On the other hand, for large values of $C$ the range of values is much
larger and the values of variables corresponding to points close to or
exactly on the target margin are tedious to adjust to the demanded
accuracy. In this case shrinking is important, and second order working
set selection is known to work best. Also, in this situation shrinking
is most endangered to make wrong decisions, so soft shrinking has an
advantage. Only the magnitude of the speed-up is really surprising.

The forest cover data is an exception to the above scheme. Here the
original algorithm is superior for all tested values of $C$, although
the difference diminishes for large $C$. Also, it seems odd that the
training times for the different target accuracies are nearly identical,
despite the fact that they can make huge differences for other data
sets. The reason for this effect is most probably the rather low number
of features~$d$: once the right weight vector is found, is it easily
tuned to nearly arbitrary precision. Also, since the data is distributed
in a rather low dimensional space there are many functionally similar
instances, which is why adaptation of frequencies of individual
variables is less meaningful than for the other problems.

Training times increase drastically for increasing values of $C$ (note
the logarithmic scale in the plots). Therefore we argue that improved
training speed is most crucial for large values of~$C$. Doubling the
training time for small values of $C$ does not pose a serious problem,
since these times are anyway short, while speeding up training for large
values of $C$ by more than an order of magnitude can make machine
training feasible in the first place. We observe this difference
directly for problems kkd-a and kkd-b.

This argument is most striking when doing model selection for $C$.
Minimization of the cross validation error is a standard method. The
parameter $C$ is usually varied on a grid on logarithmic scale. This
procedure is often a lot more compute intensive than the final machine
training with the best value of $C$, and its cost is independent of the
resulting choice of $C$. Its time complexity is proportional to the
row-wise sum of the training times in the tables, i.e., over all values
of $C$. This cost is clearly dominated by the largest tested value
(here $C = 1000$), which is where savings due to variable selection
frequencies are most pronounced.

%%%%%%%%%%%%%%%%%%%%%%%%%%%%%%%%%%%%%%%%%%%%%%%%%%%%%%%%%%%%%%%%%%%%%%%%
\section{Conclusion}
\label{sec:conclusion}

We have replaced uniform variable selection in sweeps over the data for
linear SVM training with an adaptive approach. The algorithm
extrapolates past performance into the future and turns this information
into an algorithm for adapting variable selection frequencies. At the
same time the reduction of frequencies of variables at the bounds
effectively acts as a soft shrinking technique, making explicit
shrinking heuristics superfluous. To the best of our knowledge this is
the first approach of this type for SVM training.

Our experimental results demonstrate striking success of the new method
in particular for costly cases. For most problems we achieve speed-ups
of up to an order of magnitude or even more. This is a substantial
performance gain. The speed-ups are largest when needed most, i.e., for
large training data sets and large values of the regularization
constant~$C$.

\end{document}